\documentclass[10pt]{article}
\usepackage[margin=0.74in]{geometry}
\usepackage{amsmath,amssymb,amsthm}
\usepackage{booktabs,graphicx,microtype,multicol,placeins}
\usepackage[numbers,sort&compress]{natbib}
\usepackage{xcolor,tikz}
\usetikzlibrary{positioning,arrows.meta}
\usepackage[colorlinks=true,linkcolor=blue!55!black,citecolor=blue!55!black,urlcolor=blue!55!black]{hyperref}
\hypersetup{
  pdftitle={Decoupling Is Not Identification: Supervised Evidential Learning in Next-Token Prediction},
  pdfauthor={Ge Wang},
  pdfkeywords={supervised evidential learning, next-token prediction, Dirichlet concentration, epistemic uncertainty, lexical support}
}
\newtheorem{proposition}{Proposition}
\newcommand{\pbar}{\bar p}
\newcommand{\HeldoutN}{13616}
\newcommand{\SupervisedN}{54122}
\newcommand{\HeldPairN}{204}
\newcommand{\SupPairN}{884}
\newcommand{\EntHeldPartial}{\ensuremath{0.201 \pm 0.010}}
\newcommand{\EntSupPartial}{\ensuremath{0.201 \pm 0.007}}
\newcommand{\EntHeldPair}{\ensuremath{0.822 \pm 0.021}}
\newcommand{\EntSupPair}{\ensuremath{0.888 \pm 0.011}}
\newcommand{\EntBoundary}{\ensuremath{0.772 \pm 0.003}}
\newcommand{\ImplicitPartial}{\ensuremath{0.001 \pm 0.014}}
\newcommand{\BackoffPartial}{\ensuremath{0.011 \pm 0.012}}
\newcommand{\EntropyBoundary}{\ensuremath{0.490 \pm 0.004}}
\newcommand{\KNNBoundary}{0.769}
\newcommand{\OracleBoundary}{1.000}
\newcommand{\EntNaturalRMSE}{\ensuremath{0.575 \pm 0.031}}
\newcommand{\ConstantNaturalRMSE}{\ensuremath{0.304 \pm 0.000}}
\newcommand{\EntNaturalRtwo}{\ensuremath{-2.593 \pm 0.386}}
\newcommand{\EntBalancedRMSE}{\ensuremath{1.030 \pm 0.048}}
\newcommand{\ConstantBalancedRMSE}{\ensuremath{1.673 \pm 0.000}}
\newcommand{\EntHighRMSE}{\ensuremath{0.963 \pm 0.043}}
\newcommand{\ConstantHighRMSE}{\ensuremath{1.611 \pm 0.000}}
\newcommand{\EntAURCH}{\ensuremath{0.283 \pm 0.003}}
\newcommand{\EntAURCU}{\ensuremath{0.515 \pm 0.005}}
\newcommand{\EntAURCJoint}{\ensuremath{0.283 \pm 0.003}}
\newcommand{\CEAURCPmax}{\ensuremath{0.271 \pm 0.003}}
\newcommand{\EntJointWeight}{\ensuremath{0.000 \pm 0.000}}
\newcommand{\CoupledSupPartial}{\ensuremath{-0.047 \pm 0.002}}
\newcommand{\CoupledHeldPartial}{\ensuremath{-0.036 \pm 0.007}}
\newcommand{\EntErrorSeen}{\ensuremath{0.562 \pm 0.001}}
\newcommand{\EntErrorHigh}{\ensuremath{0.119 \pm 0.001}}
\newcommand{\SuffixShortPartial}{\ensuremath{0.156 \pm 0.033}}
\newcommand{\SuffixShortN}{334}
\newcommand{\DensityNearPartial}{\ensuremath{0.208 \pm 0.014}}
\newcommand{\DensityQtwoPartial}{\ensuremath{0.159 \pm 0.012}}
\newcommand{\DensityQthreePartial}{\ensuremath{0.215 \pm 0.010}}
\newcommand{\DensityFarPartial}{\ensuremath{0.210 \pm 0.011}}

\title{Decoupling Is Not Identification:\\Supervised Evidential Learning in Next-Token Prediction}
\author{Ge Wang\\
\small Biomedical Imaging Center, Center for Biotechnology and Interdisciplinary Studies,\\[-2pt]
\small Department of Biomedical Engineering, Rensselaer Polytechnic Institute, Troy, NY, USA\\[-2pt]
\small \texttt{wangg6@rpi.edu}}
\date{August 2026}

\begin{document}
\maketitle

\begin{abstract}
A next-token probability says what a model predicts, not how much training support lies behind it. A Dirichlet head can represent this distinction by separating mean $m$ from concentration $S$, but decoupling does not identify what $S$ means. Here we propose an Evidential Next-Token Prediction (ENTOP) framework to audit this gap on character-level \emph{Moby-Dick}, using exact 8-gram count as a reproducible lexical-support label and withholding count regression from 20\% of context types. Standard implicit evidential training carries essentially no count signal beyond confidence on held-out-label types (partial Spearman $\rho=\ImplicitPartial$), whereas explicit supervision generalizes ($\rho=\EntHeldPartial$; matched-pair win $\EntHeldPair$). CE predictive entropy is at chance for unseen 8-grams (AUROC $\EntropyBoundary$), while supervised vacuity reaches $\EntBoundary$, comparable with an indexed CE-representation baseline ($\KNNBoundary$) but below the tautological corpus oracle ($\OracleBoundary$). Neither longest-suffix nor representation-distance strata explain where amortization succeeds. Increasing count weight under the digamma objective improves support fit only by sacrificing prediction. A constant predictor wins natural log-RMSE, and vacuity does not improve error deferral. These results motivate a minimum evidence protocol---confidence control, matched pairs, held-out labels, a constant baseline, and a decision test---and show that concentration can pass identification while failing calibration and utility.
\end{abstract}


\section{Introduction}

Softmax entropy cannot distinguish a sharp prediction repeatedly confirmed in training from an equally sharp extrapolation. Evidential deep learning instead emits a distribution over categorical probabilities and interprets concentration as evidence~\citep{sensoy2018evidential,josang2016subjective,ulmer2023survey}. Prior Networks explicitly train distributional uncertainty, while Posterior Networks use latent density and pseudo-counts to ground it~\citep{malinin2018prior,charpentier2020posterior}. These methods underscore a general point: a second-order output is not epistemically meaningful merely because a network emits it. Loss minimization can determine concentration through parameterization and regularization rather than an identifiable target~\citep{bengs2022pitfalls}; analogous concerns arise in evidential regression~\citep{amini2020deep}.

This paper asks one question: \emph{what label, if any, makes concentration measure training support?} Natural character text retains morphology and syntax, while every exact context count remains computable. We withhold the count loss from 20\% of context \emph{types}, although those types still participate in next-character training. Success therefore cannot be direct fitting of their count labels.

The audit tests the claim from five directions. We cross prediction loss with count supervision; freeze identical confidence-matched pairs for all models; compare supervised and held-out types; test exact and suffix-smoothed support; and evaluate the unseen-context boundary. We further add two controls that materially narrow the result: constant predictors reveal scale degeneracy, and risk--coverage tests whether lexical support improves error deferral. The resulting claim is constructive but limited: explicit supervision makes $S$ rank lexical support across context types, not a generally calibrated count or task risk.

\section{Evidential Next-Token Prediction Framework}


A two-layer causal Transformer~\citep{vaswani2017attention} maps an $L$-character context $x$ to $h\in\mathbb R^{64}$. The exact mean--concentration head is
\begin{equation}
m=\operatorname{softmax}(W_mh),\quad
S=K\alpha_0+\exp(w_s^\top h+b),\quad
\alpha=Sm,\quad \pi\sim\operatorname{Dir}(\alpha),
\label{eq:head}
\end{equation}
where $K$ is the vocabulary size and $\alpha_0=0.1$ determines the minimum concentration $K\alpha_0$. Thus $\mathbb E[\pi]=m$ exactly: $S$ cannot algebraically change prediction. We clip the exponential at $10^6$, effectively unbounded here.
The common coupled head sets $\alpha_k=e_k+1$ with $e_k\ge0$~\citep{sensoy2018evidential}.
\begin{proposition}[Tight sharpness--concentration bound]
If $\alpha_k\ge1$, then $\pbar_{\max}\le1-(K-1)/S$. Equality is attained when each nonmaximal $\alpha_j=1$. Hence $\pbar_{\max}\ge1-\epsilon$ requires $S\ge(K-1)/\epsilon$.
\end{proposition}
\noindent Coupling forbids sharp-and-thin states, but also acts as an implicit regularizer: a diffuse off-support mean automatically lowers concentration. Decoupling removes both effects. It supplies representational freedom, not an epistemic semantics.


Here we propose an evidential next-token prediction (ENTOP) framework that augments next-token probability prediction with an explicit evidence variable intended to represent how strongly the prediction is supported by training data.
Let $n(x)$ count training positions preceded by $x$. We define
\begin{equation}
S^*(x)=K\alpha_0+n(x),\qquad
\mathcal L_{\rm count}=[\log S(x)-\log S^*(x)]^2.
\label{eq:count}
\end{equation}
We use ENTOP–D to denote the digamma-based evidential variant, which uses the same decoupled mean–concentration head but trains it only with the standard digamma expected cross-entropy and annealed KL regularization, without explicit supervision of S. Thus, “S” emphasizes explicit support supervision, whereas “D” denotes the digamma evidential objective.
This borrows the total concentration of a categorical--Dirichlet update, not its full posterior vector: Eq.~\eqref{eq:head} has no per-component floor. Exact count is an auditable lexical-exposure target, not semantic evidence.

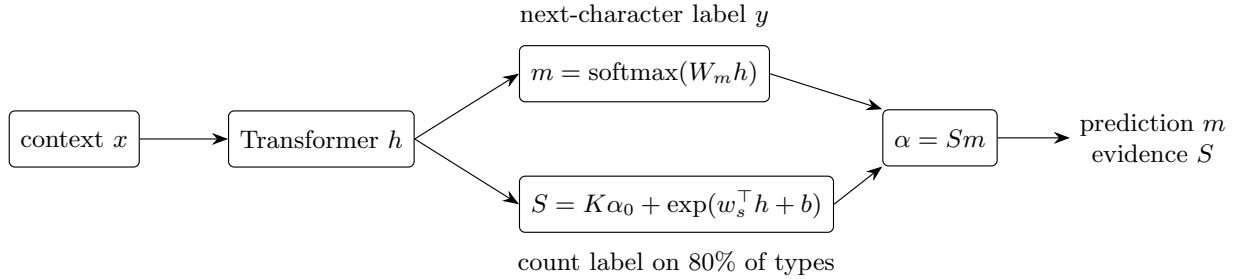
\begin{figure}[ht]
\centering
\resizebox{0.91\linewidth}{!}{%
\begin{tikzpicture}[font=\small,node distance=7mm and 11mm,
box/.style={draw,rounded corners=2pt,minimum height=7mm,inner sep=4pt},arr/.style={-{Stealth[length=2mm]}}]
\node[box] (x) {context $x$}; \node[box,right=of x] (h) {Transformer $h$};
\node[box,above right=1mm and 13mm of h] (m) {$m=\mathrm{softmax}(W_mh)$};
\node[box,below right=1mm and 13mm of h] (s) {$S=K\alpha_0+\exp(w_s^\top h+b)$};
\node[box,right=14mm of m,yshift=-8mm] (a) {$\alpha=Sm$};
\node[right=9mm of a,align=center] (o) {prediction $m$\\evidence $S$};
\draw[arr] (x)--(h); \draw[arr] (h.east)--(m.west); \draw[arr] (h.east)--(s.west);
\draw[arr] (m.east)--(a.north west); \draw[arr] (s.east)--(a.south west); \draw[arr] (a)--(o);
\node[above=1mm of m] {next-character label $y$}; \node[below=1mm of s] {count label on 80\% of types};
\end{tikzpicture}}
\caption{ENTOP--S assigns separate targets to prediction direction and evidence strength. The held-out 20\% receive language-model training but no count regression.}
\label{fig:method}
\end{figure}

We test six conditions. \textbf{CE} is a softmax baseline. \textbf{CE--S} uses Eq.~\eqref{eq:head} with cross-entropy but no loss on $S$. \textbf{ENTOP--S} adds $\beta\mathcal{L}_{\rm count}$ with $\beta=0.3$ as default. \textbf{ENTOP--D} uses the same exact head but only the standard digamma expected cross-entropy with annealed KL. \textbf{D+count} adds the count loss. \textbf{Coupled} uses $\alpha=e+1$ with digamma--KL. This grid separates count supervision from prediction objective. The \textbf{ENTOP--S} workflow is shown in Figure \ref{fig:method}.

\section{Experimental Design}

\paragraph{Corpus and split.}
We use the CC0 chapter files of \emph{Moby-Dick} distributed by stdlib~\citep{stdlibmoby}. Lowercased, ASCII-normalized chapters 1--110, 111--120, and 121--135 form train, validation, and test partitions of 1,006,806, 49,335, and 130,653 characters. With $K=48$ and $L=8$, training contains 511,526 unique context types.

\paragraph{Held-out count labels.}
Types are stratified by $\lfloor\log_2n\rfloor$ into one split shared across seeds: 80\% may receive count loss and 20\% may not. Cross-entropy samples positions from all types; count batches contain only supervised types and are balanced across count bins. Primary support analysis requires $n\ge2$ and label consistency $q(x)=\max_yc(x,y)/n(x)\ge0.9$, giving \SupervisedN{} supervised and \HeldoutN{} held-out-label types. This filter targets repeated, label-consistent exposure and excludes many high-entropy contexts.

\paragraph{Training and comparisons.}
Every condition uses a width-64, two-layer, four-head Transformer; Adam at $10^{-3}$; batch 1,024; five epochs; 250,000 positions per epoch; and five seeds. Count models receive a 512-type auxiliary batch. A post-review D+count diagnostic sweeps $\beta\in\{0.1,0.3,1,3,10\}$ on seed 0. Earlier versions used a different additive-floor head and sampling protocol; their ENTOP--D accuracy is therefore not directly comparable.

\paragraph{Support metrics.}
We report partial Spearman correlation after residualizing ranks against $H(m)$ and $m_{\max}$. Two CE-mined pair sets share the same predicted character, $|\Delta m_{\max}|<0.01$, $|\Delta H|<0.05$, $n_{\rm low}\le2$, and $n_{\rm high}\ge20$. They contain \SupPairN{} supervised and \HeldPairN{} held-out pairs; win rate is $\Pr(S_{\rm high}>S_{\rm low})$.

For scale, a constant baseline predicts the mean supervised $\log S^*$. We report natural type-weighted log-RMSE, count-bin-balanced log-RMSE, and RMSE on $n\ge20$. These answer different questions because low counts dominate the natural type distribution.

\paragraph{Boundary, interpolation, and decision utility.}
An 8-gram is unseen when $n_{\rm train}(x)=0$. We compare entropy and vacuity $u=K/S$ with two indexed references: longest matched suffix is an exact corpus-query oracle, while approximate 1-NN uses a fixed 50,000-type bank in the seed-0 CE hidden space after a 12-dimensional PCA (76.5\% variance). The oracle must attain 1.000 because full suffix length eight is equivalent to ``seen.'' It is an upper bound requiring the corpus index, not a learned predictor.

To probe interpolation, held-out types are stratified by the longest proper suffix appearing in the supervised type set, separating the $\ell\le3$ tail from $\ell=4$. We independently divide the same types into quartiles of 1-NN distance to the fixed CE-$h$ bank. The former tests lexical overlap; the latter tests whether representation proximity explains count-label amortization. We also retain a standardized suffix-backoff sweep for ENTOP--D. Finally, test errors are sorted from low to high uncertainty and
\begin{equation}
\operatorname{AURC}(r)=\frac1N\sum_{c=1}^N\frac1c\sum_{i=1}^c \mathbf1[\hat y_{(i,r)}\ne y_{(i,r)}],
\label{eq:aurc}
\end{equation}
where $(i,r)$ denotes the $i$-$th$ test example after sorting by uncertainty score r from low to high. A joint score $(1-w)z_H+wz_u$ selects $w\in\{0,.25,.5,.75,1\}$ on validation only.

\section{Results}

\begin{table*}[t]
\centering\scriptsize\setlength{\tabcolsep}{2.55pt}
\caption{Five-seed mean $\pm$ sample SD. sup/held denote supervised/held-out types. $\rho$ denotes partial Spearman correlation. Pair sets are fixed across models. Boundary is vacuity AUROC for unseen test 8-grams.}
\label{tab:main}
\resizebox{\linewidth}{!}{\begin{tabular}{lcccccccccc}
\toprule
Model & loss & head & count & Acc. & NLL & sup-$\rho$ & held-$\rho$ & sup-pair & held-pair & boundary \\
\midrule
CE & CE & -- & -- & \ensuremath{0.487 \pm 0.002} & \ensuremath{1.687 \pm 0.007} & -- & -- & -- & -- & -- \\
CE--S & CE & exact & no & \ensuremath{0.487 \pm 0.004} & \ensuremath{1.688 \pm 0.007} & \ensuremath{0.000 \pm 0.013} & \ensuremath{0.001 \pm 0.025} & \ensuremath{0.506 \pm 0.014} & \ensuremath{0.515 \pm 0.033} & \ensuremath{0.507 \pm 0.036} \\
\textbf{ENTOP--S} & CE & exact & yes & \ensuremath{0.483 \pm 0.002} & \ensuremath{1.714 \pm 0.007} & \ensuremath{0.201 \pm 0.007} & \ensuremath{0.201 \pm 0.010} & \ensuremath{0.888 \pm 0.011} & \ensuremath{0.822 \pm 0.021} & \ensuremath{0.772 \pm 0.003} \\
ENTOP--D & Digamma+KL & exact & no & \ensuremath{0.445 \pm 0.015} & \ensuremath{2.631 \pm 0.053} & \ensuremath{0.003 \pm 0.016} & \ensuremath{0.001 \pm 0.014} & \ensuremath{0.545 \pm 0.019} & \ensuremath{0.534 \pm 0.021} & \ensuremath{0.508 \pm 0.011} \\
D+count & Digamma+KL & exact & yes & \ensuremath{0.441 \pm 0.007} & \ensuremath{2.719 \pm 0.031} & \ensuremath{0.014 \pm 0.018} & \ensuremath{0.022 \pm 0.017} & \ensuremath{0.595 \pm 0.025} & \ensuremath{0.583 \pm 0.048} & \ensuremath{0.562 \pm 0.010} \\
Coupled & Digamma+KL & coupled & no & \ensuremath{0.489 \pm 0.002} & \ensuremath{2.554 \pm 0.009} & \ensuremath{-0.047 \pm 0.002} & \ensuremath{-0.036 \pm 0.007} & \ensuremath{0.555 \pm 0.009} & \ensuremath{0.553 \pm 0.021} & \ensuremath{0.522 \pm 0.005} \\
\bottomrule

\end{tabular}}
\end{table*}

\begin{figure}[t]
\centering\includegraphics[width=\linewidth]{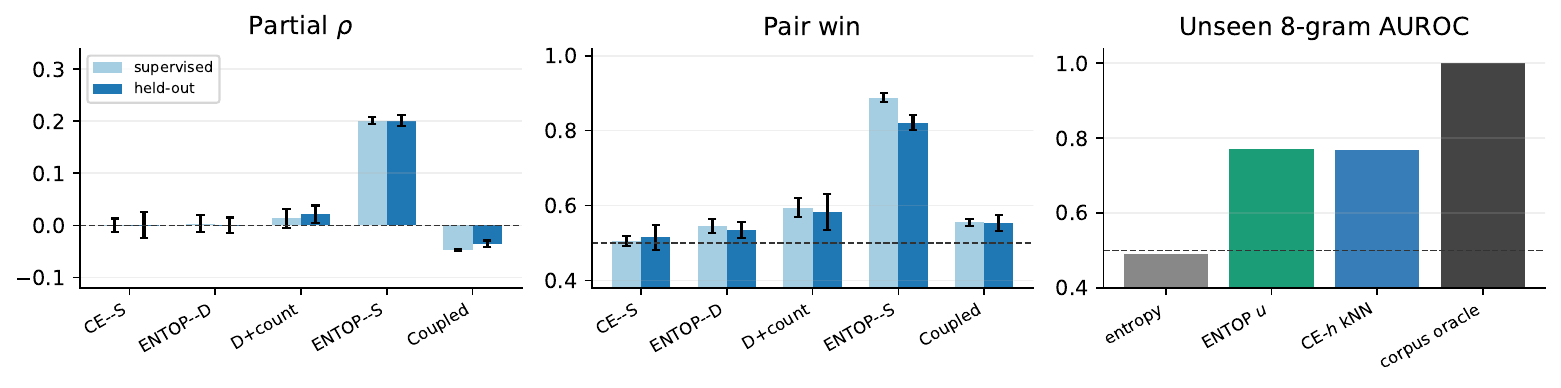}
\caption{Support identification and boundary detection. Error bars show sample SD over five seeds. The corpus oracle is exact by construction; the CE-$h$ comparator stores a 50,000-type external bank.}
\label{fig:main}
\end{figure}

\paragraph{Supervision generalizes across types.}
ENTOP--S reaches supervised/held-out partial correlations \EntSupPartial{}/\EntHeldPartial{} and pair wins \EntSupPair{}/\EntHeldPair{} (Table~\ref{tab:main}; Fig.~\ref{fig:main}). The equal global correlations are consistent with a shared amortized readout and the count-stratified split. Although the 65-parameter scalar head discourages a direct per-type table, the jointly trained backbone could still encode type-specific information; the held-out test is therefore meaningful but not absolute, while the pair gap indicates less complete transfer of local high-versus-low ordering. ENTOP--D remains null on held-out labels (\ImplicitPartial); after selecting suffix-backoff strength on supervised types, its held-out result remains \BackoffPartial{}. Coupled yields reproducibly negative residual correlations (\CoupledSupPartial{}/\CoupledHeldPartial{}). This is qualitatively consistent with Proposition~1: after mean sharpness is removed, a coupled head has no independent support degree of freedom and its residual ordering can invert, although the bound alone does not determine the sign.

\begin{figure}[t]
\centering\includegraphics[width=0.92\linewidth]{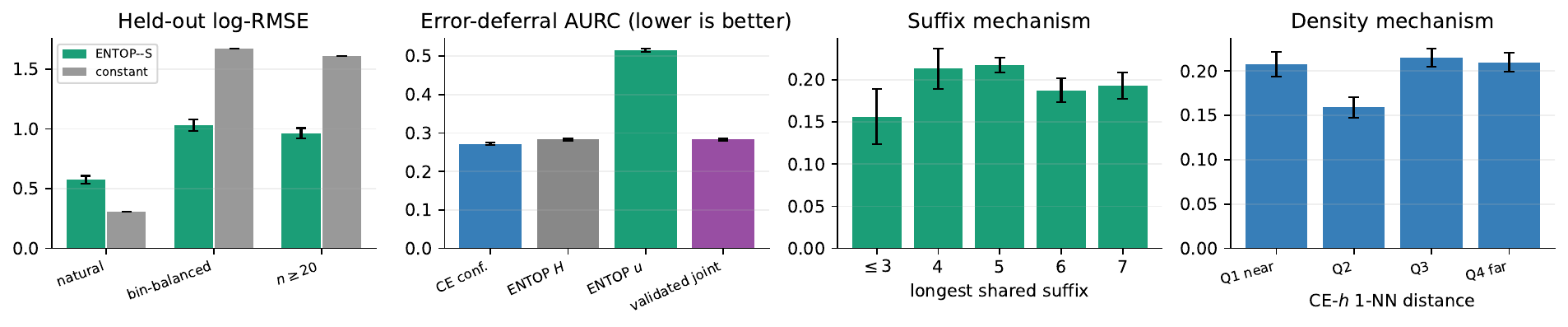}
\caption{Claim boundaries. (a) Natural RMSE favors a constant, while balanced and high-count evaluations favor ENTOP--S. (b) Vacuity fails selective error prediction. (c--d) Neither suffix length nor CE-$h$ proximity yields a monotonic support-association gradient.}
\label{fig:diag}
\end{figure}

Neither candidate interpolation mechanism explains the held-out signal (Fig.~\ref{fig:diag}c--d). The newly isolated $\ell\le3$ suffix tail contains \SuffixShortN{} types and has weaker partial correlation (\SuffixShortPartial{}), but strata $4$--$7$ are not graded. CE-$h$ proximity is also nonmonotonic: nearest-to-farthest quartiles yield \DensityNearPartial{}, \DensityQtwoPartial{}, \DensityQthreePartial{}, and \DensityFarPartial{}. Thus suffix overlap may supply a threshold effect, but neither suffix length nor fixed-bank representation distance identifies the generalization mechanism. This null is proxy-limited: distance uses seed-0 CE hidden states compressed to 12 PCA dimensions (76.5\% variance), rather than each ENTOP--S model's full representation, and quartiles may hide nonlinear tail effects.

\paragraph{Natural RMSE rewards a degenerate constant.}
On the naturally imbalanced held-out types, the constant predictor beats ENTOP--S: \ConstantNaturalRMSE{} versus \EntNaturalRMSE{} (Table~\ref{tab:scale}), giving $R^2=\EntNaturalRtwo$ relative to that constant. This does not contradict the rank results; most types have small $n$, so a constant near their center minimizes average squared error while carrying no ordering information. When count bins receive equal weight, ENTOP--S improves from \ConstantBalancedRMSE{} to \EntBalancedRMSE{}; on $n\ge20$, it improves from \ConstantHighRMSE{} to \EntHighRMSE{}. Thus supervision learns support separation, especially in the sparse high-count tail, but does not calibrate the type-majority scale.
CE--S's lower natural RMSE should likewise not be read as count recovery: its held-out partial correlation is near zero and its pair win is at chance, consistent with a near-constant concentration. At the fixed $\beta=0.3$ operating point, D+count gives the best $n\ge20$ RMSE in Table~\ref{tab:scale} (0.697) but only 0.441 accuracy. This is distinct from the seed-0 $\beta=10$ trace in Table~\ref{tab:beta}, which reports no RMSE and reaches 0.336 accuracy.

\begin{table}[t]
\centering\scriptsize\setlength{\tabcolsep}{3.2pt}
\caption{Held-out log-RMSE. ``Natural'' weights every type equally; ``balanced'' weights log-count bins equally. Each model is compared with the same constant predictor; bold model values beat that constant.}
\label{tab:scale}
\begin{tabular}{lcccccc}
\toprule
& \multicolumn{2}{c}{natural} & \multicolumn{2}{c}{bin-balanced} & \multicolumn{2}{c}{$n\ge20$}\\
Model & model & const. & model & const. & model & const.\\
\midrule
CE--S & \ensuremath{0.450 \pm 0.037} & \ensuremath{0.304 \pm 0.000} & \ensuremath{1.835 \pm 0.112} & \ensuremath{1.673 \pm 0.000} & \ensuremath{1.830 \pm 0.095} & \ensuremath{1.611 \pm 0.000} \\
ENTOP--S & \ensuremath{0.575 \pm 0.031} & \ensuremath{0.304 \pm 0.000} & \ensuremath{\mathbf{1.030 \pm 0.048}} & \ensuremath{1.673 \pm 0.000} & \ensuremath{\mathbf{0.963 \pm 0.043}} & \ensuremath{1.611 \pm 0.000} \\
ENTOP--D & \ensuremath{2.332 \pm 0.037} & \ensuremath{0.304 \pm 0.000} & \ensuremath{\mathbf{1.568 \pm 0.038}} & \ensuremath{1.673 \pm 0.000} & \ensuremath{\mathbf{1.036 \pm 0.069}} & \ensuremath{1.611 \pm 0.000} \\
D+count & \ensuremath{2.015 \pm 0.022} & \ensuremath{0.304 \pm 0.000} & \ensuremath{\mathbf{1.293 \pm 0.021}} & \ensuremath{1.673 \pm 0.000} & \ensuremath{\mathbf{0.697 \pm 0.027}} & \ensuremath{1.611 \pm 0.000} \\
Coupled & \ensuremath{2.162 \pm 0.003} & \ensuremath{0.304 \pm 0.000} & \ensuremath{\mathbf{1.394 \pm 0.001}} & \ensuremath{1.673 \pm 0.000} & \ensuremath{\mathbf{0.799 \pm 0.004}} & \ensuremath{1.611 \pm 0.000} \\
\bottomrule

\end{tabular}
\end{table}

\paragraph{The boundary is learnable, but an index is stronger.}
Entropy is at chance for the unseen 8-gram boundary (\EntropyBoundary), demonstrating that seen and unseen natural contexts can be equally predictable. ENTOP--S vacuity reaches \EntBoundary, close to the fixed 50,000-type CE-hidden kNN index (\KNNBoundary) but below the tautological exact corpus-query oracle (\OracleBoundary). The kNN comparator adds at least 600,000 stored PCA floats (50,000$\times$12), index overhead, and a query per token; ENTOP--S adds only $w_s,b$ (65 scalars) and one dot product beyond the shared backbone. This is an amortization advantage, not a superior support oracle. Moreover, the flat distance-stratified correlations show that the similar 0.772/0.769 boundary AUROCs do not establish a common mechanism.

\paragraph{Digamma loss trades prediction for count fit.}
The $\beta$ sweep rejects the claim that the digamma objective absolutely prevents an explicit target (Table~\ref{tab:beta}). Larger count weight raises held-out $\rho$ and pair win, but sharply degrades prediction: at $\beta=10$, held-out $\rho=0.133$ and pair win is 0.809, yet boundary AUROC remains 0.711 versus ENTOP--S's 0.772 and accuracy is 0.336 versus 0.483. The fixed-$\beta$ 2$\times$2 cell therefore compares operating points, not objectives in isolation; it exposes objective competition and undertraining of the predictive backbone, not an impossibility theorem.

\begin{table}[t]
\centering\scriptsize\setlength{\tabcolsep}{4pt}
\caption{Seed-0 diagnostic D+count sweep. It is a trade-off trace, not a five-seed model comparison.}
\label{tab:beta}
\begin{tabular}{ccccccc}
\toprule
$\beta$ & Acc. & NLL & sup-$\rho$ & held-$\rho$ & held-pair & boundary\\
\midrule
0.1 & 0.432 & 2.699 & 0.028 & 0.017 & 0.588 & 0.553 \\
0.3 & 0.429 & 2.769 & 0.029 & 0.038 & 0.520 & 0.548 \\
1 & 0.415 & 2.900 & 0.033 & 0.040 & 0.593 & 0.597 \\
3 & 0.387 & 3.105 & 0.069 & 0.087 & 0.706 & 0.640 \\
10 & 0.336 & 3.315 & 0.132 & 0.133 & 0.809 & 0.711 \\
\bottomrule

\end{tabular}
\end{table}

\paragraph{Lexical support does not improve error deferral.}
ENTOP--S entropy has AURC \EntAURCH{}, but vacuity has \EntAURCU{} (lower is better; Fig.~\ref{fig:diag}b). Validation chooses $w=\EntJointWeight$ for the joint score, yielding \EntAURCJoint{}---effectively entropy alone. CE confidence reaches \CEAURCPmax{}; part of its small advantage over ENTOP--S entropy reflects ENTOP--S's slightly worse accuracy and NLL under multi-task training. That cannot explain vacuity's much larger deficit. Indeed, \EntErrorSeen{} of ENTOP--S errors occur on seen 8-grams, while only \EntErrorHigh{} occur at $n\ge20$: errors mix unseen extrapolation with ambiguous seen contexts, neither of which is ordered monotonically by exposure. Thus \emph{evidence of exposure} is structurally different from \emph{evidence of correctness}.

\FloatBarrier
\section{Discussion and Conclusion}

The experiment separates representation, identification, calibration, and utility. A coupled $\alpha=e+1$ head cannot express every sharp-and-thin state; an exact $\alpha=Sm$ head removes that restriction, but neither tells the optimizer what $S$ should mean. Count labels make $S$ rank lexical exposure and recognize an unseen-string boundary, not guarantee calibrated count scale or useful error deferral. The amortization mechanism remains unresolved because neither lexical-overlap strata nor the tested CE-$h$ proximity proxy produces a monotonic gradient.

An earlier zero-structure pilot showed a strong raw correlation, which collapsed to approximately zero after adding shared linguistic structure and confidence control, suggesting that the pilot primarily learned exposure frequency because exposure was essentially the only available structure. The five present controls---confidence-adjusted correlation, matched pairs, held-out evidence labels, a constant baseline, and a decision test---form a minimum protocol for epistemic-concentration claims. Scope remains narrow: one novel, one character tokenizer, five seeds, and a filtered label-consistent subset. Exact count is convenient rather than uniquely decision-relevant; future targets should be chosen for a decision---retrieval provenance, semantic-neighborhood density, or selective risk---and evaluated by it.


In conclusion, we have proposed the ENTOP framework in which decoupling can represent sharp predictions with little evidence but cannot generically identify concentration. Explicit labels amortize lexical support across types, approaching an indexed baseline without retaining its bank. Constant and risk--coverage controls show why the label must be named precisely: support ranking is neither calibrated count nor prediction risk. Evidence-native modeling requires both a suitable representation and an explicit answer to \emph{evidence of what?}

\paragraph{Author--AI Collaboration.}
Generative AI tools, ChatGPT and Claude, were used to assist the author in brainstorming, formulation, simulation, analysis, drafting, checking, and polishing. 
The author conceptualized the ENTOP framework, collaborated with the two AI models, and takes responsibility for the content. 

\begingroup\fontsize{6.2}{6.7}\selectfont\setlength{\bibsep}{0pt}\setlength{\premulticols}{0pt}\bibliographystyle{plainnat}
\begin{multicols}{2}\bibliography{refs}\end{multicols}\endgroup
\end{document}